\documentclass[11pt,a4paper]{article}
\usepackage[hyperref]{naaclhlt2019}
\usepackage{times}
\usepackage{amsmath,amssymb}
\usepackage{graphicx,subcaption}
\usepackage{textcomp}
\usepackage{xcolor}

\usepackage{url}

\aclfinalcopy 

\newcommand{\argmin}{\operatornamewithlimits{argmin}}
\DeclareMathOperator{\E}{\mathbb{E}}
\newcommand\numberthis{\addtocounter{equation}{1}\tag{\theequation}}

\title{Riemannian Normalizing Flow on Variational Wasserstein Autoencoder for Text Modeling}

\author{
  Prince Zizhuang Wang\\
  Department of Computer Science\\
  University of California, Santa Barbara\\
  {\tt zizhuang\textunderscore wang@ucsb.edu} \\\And
  William Yang Wang\\
  Department of Computer Science\\
  University of California, Santa Barbara\\
  {\tt william@cs.ucsb.edu} \\}

\date{}

\begin{document}
\maketitle
\begin{abstract}
  Recurrent Variational Autoencoder has been widely used for language modeling and text generation tasks. These models often face a difficult optimization problem, also known as the Kullback-Leibler (KL) term vanishing issue, where the posterior easily collapses to the prior, and the model will ignore latent codes in generative tasks. 
   To address this problem, we introduce an improved Wasserstein Variational Autoencoder (WAE) with Riemannian Normalizing Flow (RNF) for text modeling.
   The RNF transforms a latent variable into a space that respects the geometric characteristics of input space, which makes posterior impossible to collapse to the non-informative prior. The Wasserstein objective minimizes the distance between the marginal distribution and the prior directly, and therefore does not force the posterior to match the prior. Empirical experiments show that our model avoids KL vanishing over a range of datasets and has better performances in tasks such as language modeling, likelihood approximation, and text generation. Through a series of experiments and analysis over latent space, we show that our model learns latent distributions that respect latent space geometry and is able to generate sentences that are more diverse. \footnote{Code could be found at \url{https://github.com/kingofspace0wzz/wae-rnf-lm}} 
\end{abstract}

\section{Introduction}
Variational Autocoder (VAE)~\cite{kingma2013auto, rezende:2015} is a probabilistic generative model shown to be successful over a wide range of tasks such as image generation~\cite{gregor2015draw, yan2016attribute2image}, dialogue generation~\cite{zhao2017learning}, transfer learning~\cite{shen2017style}, and classification~\cite{jang2016categorical}. The encoder-decoder architecture of VAE allows it to learn a continuous space of latent representations from high-dimensional data input and makes sampling procedure from such latent space very straightforward. Recent studies also show that VAE learns meaningful representations that encode non-trivial information from input~\cite{gao2018auto, zhao2017infovae}.

Applications of VAE in tasks of Natural Language Processing~\cite{bowman2015generating, zhao2017learning, zhao2018unsupervised, miao2016neural} is not as successful as those in Computer Vision. With \textit{long-short-term-memory} network (LSTM)~\cite{hochreiter1997long} used as encoder-decoder model, the recurrent variational autoencoder ~\cite{bowman2015generating} is the first approach that applies VAE to language modeling tasks. They observe that LSTM decoder in VAE often generates texts without making use of latent representations, rendering the learned codes as useless. This phenomenon is caused by an optimization problem called KL-divergence vanishing when training VAE for text data, where the KL-divergence term in VAE objective collapses to zero. This makes the learned representations meaningless as zero KL-divergence indicates that the latent codes are independent of input texts.

Many recent studies are proposed to address this key issue. ~\citet{yang2017improved, semeniuta2017hybrid} use \textit{convolutional neural network} as decoder architecture to limit the expressiveness of decoder model. ~\citet{xu2018spherical, zhao2017infovae, zhao2017learning, zhao2018unsupervised} seek to learn different latent space and modify the learning objective. And, even though not designed to tackle KL vanishing at the beginning, recent studies on \textit{Normalizing Flows}~\cite{rezende:2015, van2018sylvester} learn meaningful latent space as it helps to transform an over-simplified latent distribution into more flexible distributions. 

In this paper, we propose a new type of flow, called Riemannian Normalizing Flow (RNF), together with the recently developed Wasserstein objective~\cite{tolstikhin2017wasserstein, arjovsky2017wasserstein}, to ensure VAE models more robust against the KL vanishing problem. As further explained in later sections, the Wasserstein objective helps to alleviate KL vanishing as it only minimizes the distance between latent marginal distribution and the prior. 
Moreover, we suspect that the problem also comes from the over-simplified prior assumption about latent space. In most cases, the prior is assumed to be a standard Gaussian, and the posterior is assumed to be a diagonal Gaussian for computational efficiency. These assumptions, however, are not suitable to encode intrinsic characteristics of input into latent codes as in reality the latent space is likely to be far more complex than a diagonal Gaussian. 

The RNF model we proposed in this paper thus helps the situation by encouraging the model to learn a latent space that encodes some geometric properties of input space with a well-defined geometric metric called Riemannian metric tensor. This renders the KL vanishing problem as impossible since a latent distribution that respects input space geometry would only collapse to a standard Gaussian when the input also follows a standard Gaussian, which is never the case for texts and sentences datasets.
We then empirically evaluate our RNF Variational Wasserstein Autoencoder on standard language modeling datasets and show that our model has achieved state-of-the-art performances.
Our major contributions can be summarized as the following:
\begin{itemize}
  \item We propose Riemannian Normalizing Flow, a new type of flow that uses the Riemannian metric to encourage latent codes to respect geometric characteristics of input space.
  \item We introduce a new Wasserstein objective for text modeling, which alleviates KL divergence term vanishing issue, and makes the computation of normalizing flow easier.
  
  \item Empirical studies show that our model produces state-of-the-art results in language modeling and is able to generate meaningful text sentences that are more diverse.
\end{itemize}

\section{Related Work}
\subsection{Variational Autoencoder}
Given a set of data $\bold{x} = (x_1, x_2,...,x_n)$, a Variational Autoencoder~\cite{kingma2013auto} aims at learning a continuous latent variable that maximizes the log-likelihood $\log p(\bold x) = \log \int p_\theta(\bold x| \bold z)p(\bold z)d\bold z$. Since this marginal is often intractable, a variational distribution $q_\phi(\bold z| \bold x)$ is used to approximate the true posterior distribution $p_\theta(\bold z| \bold x)$. VAE tries to maximize the following lower bound of likelihood,
\begin{align}
\mathcal{L}(\bold \theta; \bold \phi; \bold x) & = \E_{q(\bold x)}[\E_{q_\phi(\bold z| \bold x)}[\log p_\theta(\bold x| \bold z)] \\
& \qquad - KL(q_\phi(\bold z|\bold x)||p(\bold z))]
\end{align}
where $q(\bold x)$ is the empirical distribution of input, and the prior $p(\bold z)$ is often assumed to be a standard Gaussian for simplicity. The first term in the objective is the reconstruction error, and the second one is KL divergence. For modeling text sentences, ~\citet{bowman2015generating} parameterizes both the inference model $q(\bold z|\bold x)$ and the generative model $p(\bold x|\bold z)$ as LSTMs. The reparameterization trick proposed by~\citet{rezende:2015} is used to train these two models jointly.  

\subsection{KL Divergence Term Vanishing}
Since the generative model is often an LSTM that has strong expressiveness, the reconstruction term in the objective will dominate KL divergence term. In this case, the model is able to generate texts without making effective use of latent codes as the latent variable $\bold z$ becomes independent from input when KL divergence term collapses to zero. 

There are two main approaches to address this issue. One is to explore different choices of the decoder model to control the expressiveness of LSTM. ~\citet{yang2017improved} and ~\citet{semeniuta2017hybrid} use CNN as an alternative to LSTM. The dilation technique used by~\citet{yang2017improved} also helps to control the trade-off between decoder capacity and KL vanishing. The other approach is to change the form of latent distribution and to modify the training objective. ~\citet{xu2018spherical} proposes to use hyperspherical distribution and shows that the KL vanishing problem does not happen in hypersphere. The infoVAE~\cite{zhao2017infovae} argues that VAE objective is anti-informatics which encourages KL divergence to be zero. They, therefore, add a mutual information term $I(\bold z; \bold x)$ explicitly to ensure that latent variable $\bold z$ encodes non-trivial information about $\bold x$, in which case the KL would be greater than zero as $\bold z$ is no longer independent from $\bold x$. In a similar manner, ~\citet{xiao2018dirichlet} introduces a Dirichlet latent variable to force latent codes to learn useful topic information given input documents. ~\citet{he2019lagging} and~\citet{kim2018semi} achieve the current state-of-the-art in terms of sample perplexity. Moreover, the bag-of-words loss used by~\citet{zhao2017learning} also dramatically alleviates KL vanishing while not sacrificing sample quality. In Section~\ref{rnf} and Section~\ref{model}, we introduce RNF with Wasserstein objective. Our proposed model also lies in the direction which seeks to learn more flexible latent distribution, as the main advantage of RNF is to ensure a flexible latent space able to capture geometric characteristics of input space.

\section{Background} \label{rnf}
In this section, we review the basic concepts of Riemannian geometry, normalizing flow, and Wasserstein Autoencoder. We then introduce our new Riemannian normalizing flow in Section~\ref{model}.

\subsection{Riemannian Geometry Review}
Consider an input space $\mathcal{X} \subset R^D$, a d-dimensional $(d < D)$ manifold is a smooth surface of points embedded in $\mathcal{X}$. Given a manifold $\mathcal{M}$, a Riemannian manifold is a metric space $(\mathcal{M}, G)$, where $G$ is the Riemannian metric tensor that assigns an inner product to every point on the manifold. More formally, a Riemannian metric $G: \mathcal{Z} \rightarrow R^{d \times d}$ is defined as a smooth function such that for any two vectors $u, v$ in the tangent space $T_zM$ of each point $z \in \mathcal{M}$, it assigns the following inner product for $u$ and $v$,
\begin{align*}
    <u, v>_G = u^T G(z) v
    \numberthis
\end{align*}    

The Riemannian metric helps us to characterize many intrinsic properties of a manifold. Consider an arbitrary smooth curve $\gamma (t) : [a, b] \rightarrow \mathcal{M}$ on a given manifold $\mathcal{M}$ with a Riemannian metric tensor $G$, the length of this curve is given by 
\begin{align*}
\mathcal{L} (\gamma) &= \int_a^b ||\gamma'(t)||dt \\
    &= \int_a^b \sqrt{{<{\gamma'}_{t}, {\gamma'}_{t}>}_G}dt\\
    &= \int_a^b \sqrt{{\gamma'}_{t}^{T} G(\gamma_t) {\gamma'}_t}dt
    \numberthis
\end{align*}
where $\gamma'_t$ is the curve velocity and lies in the tangent space $T_{\gamma_t}M$ at point $\gamma (t)$. When the metric tensor $G$ is equal to 1 everywhere on the curve, it becomes a metric tensor on Euclidean space, where the length of curve is defined as the integral of the velocity function, $\mathcal{L}(\gamma) = \int_a^b \sqrt{{\gamma'}_{t}^{T} \gamma'_t}dt = \int_a^b \gamma'_t dt$. Given the definition of curve length, the geodesic path between any two points can be defined as the curve that minimizes the curve length $\mathcal{\gamma}$. Namely, if $\gamma_t$ is the geodesic curve connecting $\gamma(a)$ and $\gamma(b)$, then
\begin{equation}
    \gamma_t = \argmin_\gamma \mathcal{L}(\gamma)
\end{equation}

Practically, a geodesic line is often found by optimizing the following energy function,
\begin{align*}
E(\gamma) & = \frac{1}{2} \int_a^b {\gamma'}_{t}^{T} G(\gamma_t) \gamma'_t dt\\
    \gamma_t & = \argmin_\gamma E(\gamma)
    \numberthis
\end{align*}

\noindent Note that the Euclidean metric is a special case of Riemannian metric. The more general metric tensor $G$ gives us a sense of how much Riemannian geometry deviates from Euclidean geometry. 

\subsection{Review: Normalizing Flow}

The powerful inference model of VAE can approximate the true posterior distribution through variational inference. The choice of this approximated posterior is one of the major problems. For computational efficiency, a diagonal Gaussian distribution is often chosen as the form of the posterior. As the covariance matrix is always assumed to be diagonal, the posterior fails to capture dependencies among individual dimensions of latent codes. This poses a difficult problem in variational inference. As it is unlikely that the true posterior has a diagonal form, the approximated diagonal distribution is not flexible enough to match the true posterior even in asymptotic time.

A normalizing flow, developed by~\cite{rezende:2015}, is then introduced to transform a simple posterior to a more flexible distribution. Formally, a series of normalizing flows is a set of invertible, smooth transformations $f_t: R^d \rightarrow R^d$, for $t=1,...,T$, such that given a random variable $z_0$ with distribution $q(z_0)$, the resulting random variable $z^T = (f_T \circ f_{T-1} \circ ... \circ f_1)(z_0)$ has the following density function,
\begin{align*}
    q(z_T) &= q(z_0)\prod_{t=1}^T |det \frac{\partial f_t^{-1}}{\partial z_{t-1}}| 
    \numberthis
\end{align*}

Since each transformation $f_i$ for $i=1,...,T$ is invertible, its Jacobian determinant exists and can be computed. By optimizing the modified \textit{evidence lower bound} objective,
\begin{align*}
    \ln p(x) & \ge \E_{q(z_0|x)}\big[ \ln p(x|z_T) + \sum_{t=1}^T \ln |det\frac{\partial f_t}{\partial z_{t-1}}| \big]\\
    & \qquad \qquad - KL(q(z_0|x)||p(z_T))
    \numberthis
\end{align*}
the resulting latent codes $z_T$ will have a more flexible distribution.

Based on how the Jacobian-determinant is computed, there are two main families of normalizing flow~\cite{tomczak2016improving, berg2018sylvester}: \textit{general normalizing flow} and \textit{volume preserving flow}. While they both search for flexible transformation that has easy-to-compute Jacobian-determinant, the volume-preserving flow aims at finding a specific flow whose Jacobian-determinant equals 1, which simplifies the optimization problem in equation (6). Since we want a normalizing flow that not only gives flexible posterior but also able to uncover the true geometric properties of latent space, we only consider general normalizing flow whose Jacobian-determinant is not a constant as we need it to model the Riemannian metric introduced earlier. 

\subsection{Wasserstein Autoencoder}
Wasserstien distance has been brought to generative models and is shown to be successful in many image generation tasks \cite{tolstikhin2017wasserstein, arjovsky2017wasserstein, bousquet2017optimal}. Instead of maximizing the \textit{evidence lower bound} as VAE does, the Wasserstein Autoencoder \cite{tolstikhin2017wasserstein} optimizes the optimal transport cost~\cite{villani2008optimal} between the true data distribution $P_X(x)$ and the generative distribution $P_G(x)$. This leads to the Wasserstein objective,
\begin{align*}
D(P_X, P_G)& = \inf_{Q(Z|X) \in \mathcal{Q}}\E_{P_X}\E_{Q(Z|X)}[c(X, G(Z))]\\
& \qquad + \lambda D_Z(Q_Z, P_Z)
\numberthis
\end{align*}
where $c(\cdot)$ is the optimal transport cost, $G:\mathcal{Z}\rightarrow \mathcal{X}$ is any generative function, and the coefficient $\lambda$ controls the strength of regularization term $D_Z$. Given a positive-definite reproducing kernel $k: \mathcal{Z} \times \mathcal{Z} \rightarrow \mathcal{R}$, the regularization term $D_Z$ can be approximated by the \textit{Maximum Mean Discrepancy} (\textbf{MMD})~\cite{gretton2012kernel} between the prior $P_Z$ and the aggregate posterior $Q_Z(z) = \int q(z|x)p(x)dx$,
\begin{align*}
MMD_k(P_Z, Q_Z)
    & = ||\int_\mathcal{Z}k(z, \cdot)dP_Z - \int_\mathcal{Z} k(z, \cdot) dQ_Z||
    \numberthis
\end{align*}

\section{Our Approach} \label{model}

In this section we propose our \textit{Riemannian Normalizing Flow} (\textbf{RNF}). RNF is a new type of flow that makes use of the Riemannian metric tensor introduced earlier in Section~\ref{rnf}.
This metric enforces stochastic encoder to learn a richer class of approximated posterior distribution in order to follow the true geometry of latent space, which helps to avoid the local optimum in which posterior collapses to a standard prior. We then combine this with WAE and we will explain why and how WAE should be used to train with RNF.

\subsection{Riemannian Normalizing Flow}
\begin{figure}[t]
    \centering
    \includegraphics[width=0.38\textwidth]{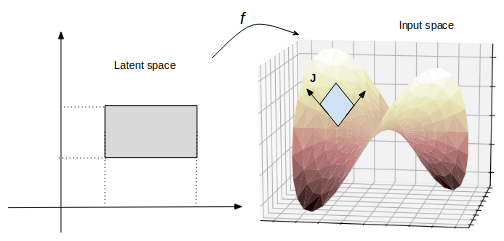}
    \caption{Parameterization of the input manifold by latent space and generative function $f$. $J$ is the Jacobian tangent space at a point on the manifold, which reflects how curved the neighborhood is around that point.}
    \label{fig:map}
    \vspace{-3ex}
\end{figure}

\begin{figure*}[t]
\centering
    \begin{subfigure}[b]{0.32\textwidth}
    \centering
        \includegraphics[width=0.7\textwidth]{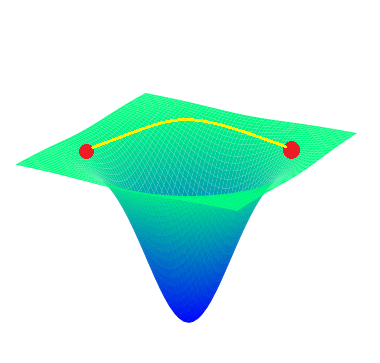}
        \label{3d}\vspace{-1ex}
    \end{subfigure}
     \begin{subfigure}[b]{0.32\textwidth}
     \centering
        \includegraphics[width=0.7\textwidth]{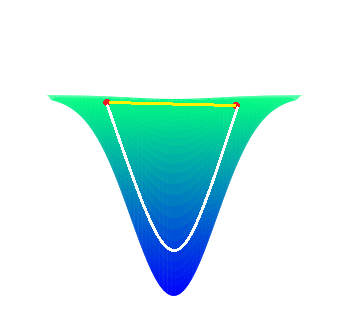}
        \label{fig:1d}\vspace{-1ex}
    \end{subfigure}
    \begin{subfigure}[b]{0.32\textwidth}
    \centering
        \includegraphics[width=0.7\textwidth]{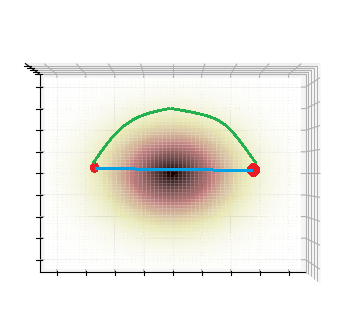}
        \label{fig:2d}\vspace{-1ex}
    \end{subfigure}
    \vspace{-2ex}
     \caption{An example when latent space does not reflect input space. \textit{Left:} a manifold that is highly curved in the central region. The yellow line is the geodesic (shortest) path connecting two sample points shown on the manifold. \textit{Right:} The projection of manifold into 2D latent space, where the color brightness indicates curvature with respect to the manifold. The green line is the geodesic path if taking the curvature into account, while the blue line is the geodesic path if we regard latent space as Euclidean. \textit{Middle:} The corresponding geodesic paths projected back from latent space to manifold. The white line corresponds to the straight geodesic path in Euclidean space. It is far longer than the true geodesic on manifold since it does not take the curvature into account in latent space.}\label{fig:curvature}
     \vspace{-2ex}
\end{figure*}

In the context of VAE, learning a latent space that is homeomorphic to input space is often very challenging. Consider a manifold $\mathcal{M} \subset R^D$, a generator model $x = f(z): \mathcal{Z} \rightarrow R^D$ serves as a low-dimensional parameterization of manifold $\mathcal{M}$ with respect to $z \in \mathcal{Z}$. For most cases, latent space $\mathcal{Z}$ is unlikely to be homeomorphic to $\mathcal{M}$, which means that there is no invertible mapping between $\mathcal{M}$ and $\mathcal{Z}$. And, since the inference model $h:\mathcal{M} \rightarrow \mathcal{Z}$ is nonlinear, the learned latent space often gives a distorted view of input space. Consider the case in Figure~\ref{fig:curvature}, where the leftmost graph is the input manifold, and the rightmost graph is the corresponding latent space with curvature reflected by brightness. Let us take two arbitrary points on the manifold and search for the geodesic path connecting these two points. If we consider the distorted latent space as Euclidean, then the geodesic path in latent space does not reflect the true shortest distance between these two points on the manifold, as a straight line in the latent space would cross the whole manifold, while the true geodesic path should circumvent this hole. This distortion is caused by the non-constant curvature of latent space. Hence, the latent space should be considered as a curved space with curvature reflected by the Riemannian metric defined locally around each point. As indicated by the brightness, we see that the central area of latent space is highly curved, and thus has higher energy. The geodesic path connecting the two latent codes minimizes the energy function $E(\gamma) = \frac{1}{2}\int {\gamma'}_{t}^{T} G(\gamma_t)\gamma'_t dt$, indicating that it should avoid those regions with high curvature $G$.   

The question now becomes how to impose this intrinsic metric and curvature into latent space. In this paper, we propose a new form of normalizing flow to incorporate with this geometric characteristic. First, consider a normalizing flow $f: \mathcal{Z} \rightarrow \mathcal{Z'}$, we can compute length of a curve in the transformed latent space $\mathcal{Z'}$,
\begin{align*}
\mathcal{L}(f(\gamma_t)) & = \int_a^b ||\bold{J}_{\gamma_t}\gamma'_t||dt\\
& =\int_a^b \sqrt{\gamma'_t \bold{J}_{\gamma_t}^T \bold{J}_{\gamma_t}\gamma'_t}dt\\
& = \int_a^b \sqrt{\gamma'_t G(\gamma_t) \gamma'_t}dt
\numberthis
\end{align*}
where 
\begin{align*}
\gamma_t & : [a, b] \rightarrow \mathcal{Z'}, \qquad a,b\in \mathcal{Z}\\
\bold{J}_{\gamma_t} & = \frac{\partial f}{\partial \bold{z}}\Bigr|_{\bold{z} = \gamma_t}\\
G(\gamma_t) & = \bold{J}_{\gamma_t}^T \bold{J}_{\gamma_t}\\
& = (\frac{\partial f}{\partial \bold{z}})^T(\frac{\partial f}{\partial \bold{z}})\Bigr|_{\bold{z} = \gamma_t}
\numberthis
\end{align*}

\noindent $\bold{J_{\gamma_t}}$ is the Jacobian matrix defined at $\gamma_t$. In our case, the Riemannian metric tensor $G$ is the inner product of Jacobian $\bold{J}_{\gamma_t}$ and is therefore symmetric positive definite. It reflects input space curvature in low-dimensional parameterization $\mathcal{Z'}$. In a highly curved region, the metric tensor $G = \bold{J}^T \bold{J}$ is larger than those in other areas, indicating that the latent representation of input manifold has lower curvature, or area of low energy, as any geodesic connecting each pair of points on the manifold favors lower energy path. This implies that those regions outside of data manifold should have high curvature reflected in their low-dimensional parameterization $\mathcal{Z'}$.

In this paper, we introduce \textit{Riemannian normalizing flow} (\textbf{RNF}) to model curvature. For simplicity, we build our model based on \textit{planar flow}~\cite{rezende:2015}. A planar flow is an invertible transformation that retracts and extends the support of original latent space with respect to a plane. Mathematically, a planar flow $f: \mathcal{Z} \rightarrow \mathcal{Z'}$ has the form,
\begin{equation}
    f(\bold{z}) = \bold{z} + \bold{u}h(\bold{w^T}\bold{z} + b)
\end{equation}
where $h: R^d \rightarrow R^d$ is any smooth non-linear function and is often chosen as $tanh(\cdot)$. The invertibiliy condition is satisfied as long as $\bold{u^T}\bold{w} \ge -1$. Its Jacobian-determinant with respect to latent codes $\bold{z}$ is very each to compute,
\begin{align*}
|det\frac{\partial f}{\partial \bold{z}}| & = |1 + \bold{u^T}\phi(\bold{z})\bold{w}|\\
\phi(\bold{z}) & = h'(\bold{w^T}z + b)
\numberthis
\end{align*}

With the Jacobian-determinant of planar flow, it is straightforward to compute the determinant of metric tensor $G$. To see that, note that since $\frac{\partial f}{\partial \bold{z}}: R^d \rightarrow R^d$ is a square matrix with full column rank due to invertibility of $f$, we have
\begin{align*}
|det\ G| & = \Bigr| \frac{\partial f}{\partial \bold{z}}^T  \Bigr| \Bigr| \frac{\partial f}{\partial \bold{z}}  \Bigr| = \Bigr| \frac{\partial f}{\partial \bold{z}}  \Bigr|^2
\numberthis
\end{align*}

To ensure well-behaved geometry in a transformed latent space, we need the Jacobian-determinant $|\frac{\partial f}{\partial \bold{z}}|$ to be large  in region with high curvature $|det\ G|$. Hence, we propose to model the metric tensor with the inverse multiquadratics kernel function $\bold{\mathcal{K}}$ used by~\cite{tolstikhin2017wasserstein} and a Gaussian kernel, that is,
\begin{align*}
    \mathcal{K}_m(\bold{z, c_k}) & = C / (C + ||\bold{z} - \bold{c_k}||_2^2)\\
    \bold{k} & = \argmin ||\bold{z} - \bold{c_k}||_2^2\\
    \mathcal{K}_g(\bold{z, c_k}) & = \exp{(-\beta_k ||\bold{z} - \bold{z_k}||_2^2)}
\end{align*}

\noindent where $\bold{c_k}, k=1,2,3,...,K$ are clusters of latent codes, and $\beta_k$ is the bandwidth. We observe that the inverse multiquadratics kernel $\mathcal{K}_m$ generally performs better. We use the above kernels as constraints over the Jacobian-determinant, so that,
\begin{align*}
|det \frac{\partial f'}{\partial \bold{z}}| = |1 + \bold{u^T(z)}\phi(\bold{z})\bold{w}|\cdot \mathcal{K}(\bold{z}, \bold{c_k})
\numberthis
\end{align*}
As we explained earlier in this section, latent representation of region outside of input manifold should have high curvature in latent space. During training, we seek to maximize this regularized Jacobian $|det \frac{\partial f'}{\partial \bold{z}}|$ rather than the original one. This ensures that those latent codes within latent clusters, and therefore very likely to be on or near input manifold in input space, have much smaller curvature $|det \ G| = |\frac{\partial f}{\partial \bold{z}}|^2$ than those outside of latent clusters, as those outside of manifold would seek larger Jacobian in order to counter-effect the regularization term $\mathcal{K}$. The latent space $\mathcal{Z}'$ transformed by normalizing flow $f$ is thus curved with respect to input manifold. This type of normalizing flow thus learns a latent space to respect geometric characteristics of input space. The KL vanishing problem is then unlikely to happen with a curved latent space. This is because most high-dimensional data in real life forms a curved manifold which is unlikely sampled from a multivariate standard Gaussian. Then, if the latent space reflects curvature of a curved manifold, the support of latent codes certainly does not follow a standard Gaussian either. This helps to push the posterior $q(\bold z|\bold x)$ away from the standard Gaussian and never collapse to a non-informative prior.

\begin{table*}[t]
\small
    \begin{center}
    \begin{tabular}{l l l l}
    \hline \bf Model & \bf PTB & \bf YAHOO & \bf YELP\\ 
     & NLL(KL) \qquad PPL & NLL(KL) \qquad PPL & NLL(KL) \qquad PPL\\
    \hline
    LSTM-LM ** & 116.2\ (-) \qquad \ \ \ 104.2 & 334.9\ (-) \qquad \quad \ \ 66.2 & - \qquad -\\
    VAE & 105.2\ (1.74) \quad \ \  121.1 & 339.2\ (0.01) \qquad 69.9 & 198.6\ (0.01) \qquad 55.0 \\
    VAE-NF & 96.8\ (0.87) \quad \ \ \ \ 82.9 & 353.8\ (0.10) \qquad 83.0 & 200.4\ (0.10) \qquad 62.5 \\
    lagging-VAE ** &  - - &\bf 326.6\ (6.70) \qquad 64.9 & --\\
    vmf-VAE & 96.0\ (5.70) \qquad 79.6 & 359.3\ (17.9) \qquad 89.9 & 198.0\ (6.40) \qquad 54.0\\
    \bf {WAE-RNF} & \bf 91.9\ (15.4) \qquad 66.1 &\bf 339.0\ (3.00) \qquad 71.6 & \bf 183.9\ (12.7) \qquad 41.1\\
    \hline
    \end{tabular}
    \end{center}
    \caption{Language Modeling results on PTB, YAHOO and YELP 13 Reviews. ** are results gathered from~\cite{yang2017improved, xiao2018dirichlet, he2019lagging}. Negative log-likelihood (NLL) is approximated by its lower bound, where the number in parentheses indicates KL-divergence. NF stands for the standard planar normalizing flow without Riemannian curvature.}
    \label{tab:language}
\end{table*}

\begin{table*}[t]
\small
    \begin{center}
    \begin{tabular}{l| l l l l| l l l l}
    \hline \bf Data & & \qquad PTB & & & & \qquad Yelp & \\
    \hline \bf Model \qquad & \bf NLL & \bf Re-KL & \bf Log-J & \bf PPL  & \bf NLL & \bf Re-KL & \bf Log-J & \bf PPL\\
    \hline 
        \bf WAE & 104.9 & --(1.9) & -- & 131. & 198.5 & --(1.9) & -- & 55\\
        \bf WAE-NF & 92.3 & 14.3 & 14.3 & 67.3 & 184.3 & 13.9 & 14.2 & 41.4 \\
        \bf WAW-RNF  &\bf 91.9 &\bf 15.4 &\bf 15.2 &\bf 66.1 &\bf 183.9 &\bf 12.7 &\bf 12.1 &\bf 41.1 \\
        \hline
    \end{tabular}
    \end{center}
    \caption{Language Modeling using WAE-RNF. We report NLL, PPL, Sum of Log Jacobian, and KL divergence between $q(\bold z'|\bold x)$ and $p(\bold z')$.}
    \label{tab:language_wae}
\end{table*}

\subsection{RNF Wasserstein Autoencoder}
Here we consider using the Wasserstein objective to model the latent marginal distribution of a curved latent space learned by an RNF. The Wasserstein objective with MMD is appealing in our case for two main reasons.

First, instead of minimizing the KL-divergence $KL(q(\bold z| \bold x)||p(\bold z))$, it minimizes distance between $q(\bold z) = \int q(\bold z|\bold x)p(\bold x)dx$ and $p(\bold z)$, which encourages the marginal distribution of latent space to be as close as the prior while not affecting individual posterior distribution $q(\bold z|\bold x)$ conditioned on each input. This makes the KL-divergence between posterior and prior, and equivalently the mutual information between latent codes and input sentences $I(\bold z,\bold x) = KL(q(\bold z,\bold x)||q(\bold z)p(\bold x)) = \E_{p(\bold x)}[KL(q(\bold z|\bold x)||p(\bold z))] - KL(q(\bold z)||p(\bold z)) $ impossible to be vanished as the objective does not require it to be small. Since the learned latent codes and input sentences have non-zero mutual information, the generative model will not ignore latent codes when generating texts. 

Second, the MMD regularization in WAE makes it possible to optimize normalizing flow without computing the Jacobian-determinant explicitly. The use of MMD is necessary as getting a closed form KL divergence is no-longer possible after we apply RNF to the posterior. And, since the generative function $G$ in the reconstruction term of Wasserstein objective can be any function or composition of functions~\cite{tolstikhin2017wasserstein}, we can easily compose an RNF function into $G$ such that the reconstructed texts are $\Bar{X} = G(f(Z)) = G(Z'), Z' \in \mathcal{Z}'$.

Now, given a series of RNF $F = f_K \circ ... \circ f_1 $, and let $\mathcal{Z}_K$ be the \textbf{curved} latent space after applying $K$ flows over the original latent space $\mathcal{Z}$, we optimize the following RNF-Wasserstein objective,
\begin{align*}
D(P_X, P_G)
& = \inf_{Q(Z|X) \in \mathcal{Q}}\E_{P_X}\E_{Q(Z|X)}[c(X, G(Z'))]\\
& \qquad + \lambda MMD(Q_{Z'}, P_{Z'})\\
& + \alpha (KL(q(\bold z|\bold x)||p(\bold z)) - \sum \log |det \frac{\partial f'}{\partial \bold z}|)
\numberthis
\end{align*}
where $Z \sim \mathcal{Z}$, $Z' \sim \mathcal{Z'}$, and $Z'=F(Z)$. We approximate MMD term with the Gaussian kernel $k(\bold z,\bold z') = e^{-||\bold z-\bold z'||^2}$, that is, $MMD(p, q) = \E_{p(\bold z),p(\bold z')}[k(\bold z,\bold z')] + \E_{q(\bold z),q(\bold z')}[k(\bold z,\bold z')] - 2\E_{p(\bold z),q(\bold z')}[k(\bold z,\bold z')]$.

Here we choose to minimize the MMD distance between the prior $P_Z$ and the marginal of non-curved latent space. This makes sampling procedure for generation tasks much easier, as it is easy to sample a latent code from a non-informative prior. We can get a sample $\bold z'$ from $\mathcal{Z'}$ indirectly by sampling: $\bold z \sim P_Z(\bold z)$ and $\bold z' \sim F(\bold z)$. 
On the other hand, it would be much more difficult to sample from a curved latent space $\mathcal{Z'}$ directly as the only prior knowledge we have about $\mathcal{Z'}$ is the curvature reflected by RNF implicitly and hence we do not know the support of $Q(Z')$. 

\section{Experimental Results}
In this section, we investigate WAE's performance with Riemannian Normalizing Flow over language and text modeling. 
\subsection{Datasets}
We use Penn Treebank~\cite{marcus1993building}, Yelp 13 reviews~\cite{xu2016cached}, as in~\cite{xu2018spherical, bowman2015generating}, and Yahoo Answers used in~\cite{xu2018spherical, yang2017improved} to follow and compare with prior studies.
We limit the maximum length of a sample from all datasets to 200 words. 
The datasets statistics is shown in Table~\ref{datasets}.

\begin{table}[h]
\small
    \begin{center}
    \begin{tabular}{l|c c c c}
    \hline  \bf Data & \bf Train & \bf Dev & \bf Test & \bf Vocab \\ \hline
        PTB & 42068 & 3370 & 3761 & 10K \\
        Yelp13 & 62522 & 7773 & 8671 & 15K\\
        Yahoo & 100K & 10K & 10K & 20K\\
    \hline
    \end{tabular}
    \end{center}
    \caption{\label{datasets} Datasets statistics; The numbers reflect size of each dataset. Vocab is the vocabulary size.}
    \vspace{-2ex}
\end{table}

\subsection{Experimental Setup}

For each model, we set the maximum vocabulary size to 20K and the maximum length of input to 200 across all data sets. Following ~\citet{bowman2015generating}, we use one-layer undirectional LSTM for both encoder-decoder models with hidden size 200. Latent codes dimension is set to 32 for all models. We share Word Embeddings of size 200. For stochastic encoders, both $MLP_\mu$ and $MLP_\sigma$ are two layer fully-connected networks with hidden size 200 and a batch normalizing output layer ~\cite{ioffe2015batch}. 

We use Adam~\cite{kingma2015adam} with learning rate set to $10^{-3}$ to train all models. Dropout is used and is set to 0.2. We train all models for 48 epochs, each of which consists of 2K steps. For models other than WAE, KL-annealing is applied and is scheduled from 0 to 1 at the 21st epoch. 

For vmf-VAE~\cite{xu2018spherical}, we set the word embedding dimension to be 512 and the hidden units to 1024 for Yahoo, and set both of them to 200 for PTB and Yelp. The temperature $\kappa$ is set to 80 and is kept constant during training.

For all WAE models, we add a small KL divergence term to control the posterior distribution. We found that if we only use RNF with MMD as the distance metric, then the posterior may diverge from the prior such that no reasonable samples can be generated from a standard Gaussian variable. Hence, for all data sets, we schedule the KL divergence weight $\alpha$ from 0 to 0.8, and the weight of the MMD term is set as $\lambda = 10 - \alpha$. $\beta_k$ of RBF is set to 10 for all models. For RNF, we use pre-trained standard VAE models to gather the clusters $\bold{c_k}$, $k=1,...,K$, of latent codes, where we set the number of clusters to be 20. We use three normalizing flow for all experiments.

\paragraph{Hyperparameter of $\mathcal{K}_m$} When using the inverse multiquadratics kernel $\mathcal{K}_m(\bold z, \bold c_k) = C / (C + ||\bold z - \bold c_k||_2^2)$ for RNF, we follow the choice of hyperparameter in~\cite{tolstikhin2017wasserstein}. We set $C = 2 \cdot d \cdot s$, where $d$ is the dimensionality of latent codes $\bold z$, and $s$ is ranged in $(0.1, 0.2, 0.5, 1, 2, 5, 10)$. The final kernel is computed by $\mathcal{K}_m(\bold z, \bold c_k) = \sum_s 2ds / (2ds + ||\bold z - \bold c_k||_2^2)$. As explained by~\cite{tolstikhin2017wasserstein}, this strategy allows us to explore a wider range of hyperparameter in one setting.

\begin{table*}[t]
\small
    \centering
    \begin{tabular}{l}
    \hline

    the company said it will be sold to the company 's promotional programs and \_UNK\\
the company also said it will sell \$  n million of soap eggs turning millions of dollars \\

the company said it will be \_UNK by the company 's \_UNK division n\\

the company said it would n't comment on the suit and its reorganization plan \\
    \hline
    this is a reflection of socialism and capitalism\\
    the company also said it will sell its \_UNK division of the company 's \_UNK\\
    earlier this year the company said it will sell \$ n billion of assets and \_UNK to the u.s \\
    last year he said the company 's earnings were n't disclosed\\
    \hline
    \\ 
    \hline

    one of my favorite places to eat at the biltmore . the food is good . and the food is good.\\
    very good food . the food was very good . the service was great and the food is very good.\\
    one of my favorite places to eat and a great breakfast spot . the food is great . the staff is friendly.\\
    took a few friends to join me to the \_UNK . i was n't sure what to expect.\\
    \hline
    one of my favorite places to eat at the biltmore . the food is good , the service was great .\\ 
 
i love the fact that they have a lot
took a few friends to join me to the \_UNK  .\\  
i have been to this location a few times , but i ' ve never been disappointed\\
let me start by saying that i love the idea of how to describe it . \\
    \hline
    
 
    \end{tabular}
    
    \caption{Qualitative comparison between VAE and our proposed approach. First row: PTB samples generated from prior $p(\bold z)$ by VAE (\textit{upper half}) and WAE-RNF (\textit{lower half}). Second row: Yelp samples generated from prior $p(\bold z)$ by VAE (\textit{upper half}) and WAE-RNF (\textit{lower half}).}
    
    \label{sample1}
\end{table*}

\subsection{Language Modeling Results}
We show the language modeling results for PTB, Yahoo and Yelp in Table~\ref{tab:language}. We compare negative log-likelihood (NLL), KL divergence, and perplexity (PPL) with all other existing methods. The negative log-likelihood is approximated by its lower bound.

We use the negative of ELBO to approximate NLL for all VAE models. For those with normalizing flows, we use the modified ELBO, which is $\mathcal{L} = \E_{q(\bold z^{(0)}|\bold x)}[\log p(\bold x|\bold z^{(T)}) - \log q(\bold z^{(0)}|\bold x) + \log p(\bold z^{(T)})] + \E_{q(\bold z^{(T)})}[\sum_{t=1}^T log|\frac{\partial f^{(t)}}{\partial \bold z^{(t-1)}}|]$. 

The numbers show that KL-annealing and dropout used by~\citet{bowman2015generating} are helpful for PTB, but for complex datasets such as Yahoo and Yelp, the KL divergence still drops to zero due to the over-expressiveness of LSTM. This phenomenon is not alleviated by applying normalizing flow to make the posterior more flexible, as shown in the third row. Part of the reason may be that a simple NF such as a planar flow is not flexible enough and is still dominated by a powerful LSTM decoder.

We find that the KL vanishing is alleviated a little bit if using WAE, which should be the case as WAE objective does not require small KL. We also find that simply applying a planar flow over WAE does not improve the performance that much. On the other hand, using RNF to train WAE dramatically helps the situation which achieves the lowest text perplexity on most conditions except for YAHOO Answers, where~\cite{he2019lagging, yang2017improved, kim2018semi} have the current state-of-the-art results. We want to emphasize that CNN-VAE~\cite{yang2017improved} and SA-VAE~\cite{kim2018semi} are not directly comparable with other current approaches. Here, we compare with models that use LSTM as encoder-decoder and have similar time complexity, while the use of CNN as decoder in CNN-VAE would dramatically change the model expressiveness, and it is known that SA-VAE's time complexity~\cite{kim2018semi, he2019lagging} is much higher than all other existing approaches. 

\subsection{How Good is Riemannian Latent Representation?}

\paragraph{Mutual information between $\mathcal{Z'}$ and $\mathcal{X}$}


\begin{figure}
\centering
    \includegraphics[width=.35\textwidth]{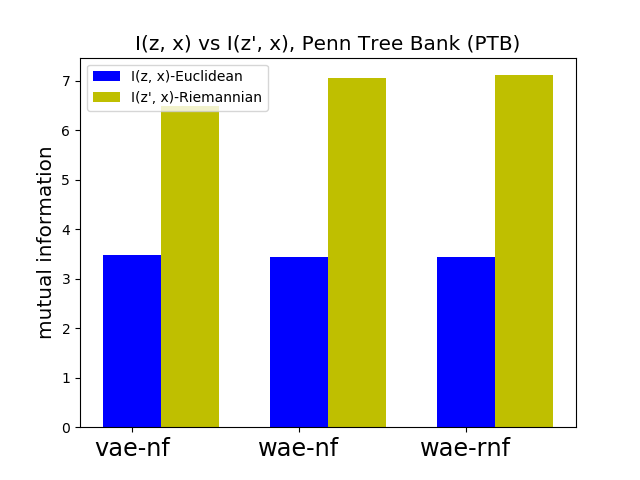}
    \caption{PTB. Comparison between the amount of mutual information stored in latent codes for different models.}
    \label{fig:mi1}
\end{figure}
\begin{figure}
\centering
    \includegraphics[width=.35\textwidth]{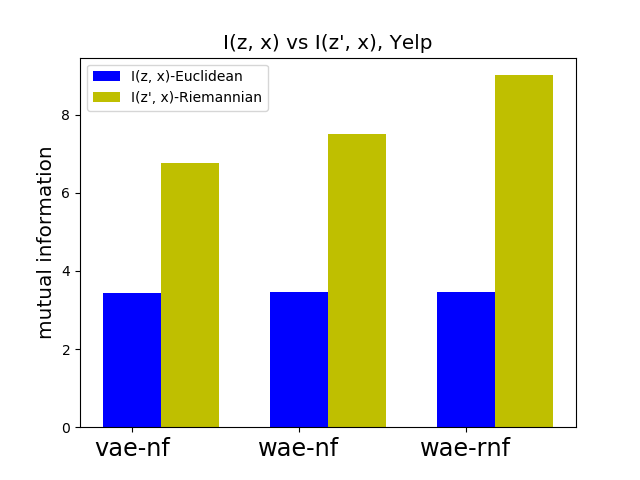}
    \caption{Yelp. Comparison between the amount of mutual information stored in latent codes for different models.}
    \label{fig:mi2}
\end{figure}

   

One important question is how useful are latent variables. Since no metrics are perfect~\cite{P18-1083}, we should not just look at sample perplexity to judge how good a latent code is. Hence, we also investigate how much information can be encoded into latent codes. We believe that the mutual information term $I(\bold z; \bold x)$ is a better metric regarding the usefulness of latent codes than sample perplexity, as it tells us directly how much information we can infer from $\bold x$ by looking $\bold z$. 

We use Monte Carlo method~\cite{metropolis1949monte} to get an approximation of $I(\bold z,\bold x) = KL(q(\bold z,\bold x)||q(\bold z)p(\bold x)) = \E_{p(\bold x)}[KL(q(\bold z|\bold x)||p(\bold z))] - KL(q(\bold z)||p(\bold z)) $. 
We compared mutual information between input $\bold x$ and latent codes $\bold z$ sampled from Euclidean latent space $\mathcal{Z}$ and Riemannian latent space $\mathcal{Z'}$ respectively. We see that even though NF does not necessarily help WAE to achieve the lowest perplexity, it does make latent codes to preserve more information about the input. 
For WAE trained with RNF, sample perplexity and mutual information metric are both good. It is cleary that $I(\bold z', \bold x) > I(\bold z, \bold x)$, where $\bold z'$ is sampled from the curved space, and $\bold z$ is the sample transformed by the normal planar flow. This further strengthens our confidence over the usefulness of the curved latent space $\mathcal{Z'}$.

\paragraph{Generating Texts from latent spaces}
Another way to explore latent space is to look at the quality of generated texts. Here we compare sentences generated from methods that do not use Wasserstein objective and RNF with those generated from curved latent space $\mathcal{Z'}$ learned by WAE. 

We observe that texts generated from flat Euclidean space are not as diverse as the ones generated from curved space learned by WAE-RNF. This is largely related to the nature of Wasserstein objective. In WAE, the KL-divergence $KL(q||p))$ between the prior and the posterior $q(\bold z|\bold x)$ conditioned on each input $\bold x$ does not need to be small to optimize the Wasserstein objective. This indicates that the marginal $q(\bold z)$ is able to match to the prior $p(\bold z)$ while allowing each posterior $q(\bold z|\bold x)$ to have a much more diverse support than that of a standard Gaussian $p(\bold z)$. Therefore, if we randomly generate samples from curved latent space $\mathcal{Z'}$ many times, we are likely to get samples scattered in different support of distinct posterior conditioned on different input $\bold x$. Hence, the reconstructed sentences will have a much more diverse meaning or structure. 

\section{Conclusion}
In this paper, we introduced Riemannian Normalizing Flow to train Wasserstein Autoencoder for text modeling. This new model encourages learned latent representation of texts to respect geometric characteristics of input sentences space. Our results show that 
RNF WAE does significantly improve the language modeling results by modeling the Riemannian geometric space via normalizing flow.

\section*{Acknowledgement}
We want to thank College of Creative Studies and Gene $\&$ Lucas Undergraduate Research Fund for providing scholarships and research opportunities for Prince Zizhuang Wang. We also want to thank Yanxin Feng from Wuhan University for helpful discussion about Riemannian Geometry, and Junxian He (CMU), Wenhu Chen (UCSB), and Yijun Xiao (UCSB) for their comments which helped us improve our paper and experiments.

\bibliographystyle{acl_natbib}

\appendix

\end{document}